\documentclass{article}
\usepackage{conll97}

\begin{document}
\title{Explanation-Based Learning of Data-Oriented Parsing}
\author{K. Sima'an\\
        Research Institute for Language and Speech, \\ Utrecht University,\\
        Trans 10, 3512 JK Utrecht, The Netherlands\\
         khalil.simaan@let.ruu.nl}
\date{}
\bibliographystyle{apalike}
\maketitle
\input epsf
\newcommand{\citeauthoryear}{}
\newcommand{\TB}{\cal TB}
\newcommand{\hs}{\hspace*{.1cm}}
\newcommand{\lra}{$\rightarrow$}
\newcommand{\Corp}{$\cal C$\hs}
\newcommand{\VN}{$V_{N}$\/}
\newcommand{\VT}{$V_{T}$\/}
\newcommand{\TSG}{(\VN,~\VT,~$S$,~\Corp)\/} 
\newcommand{\STSG}{\mbox{(\VN, \VT, $S$, ${\cal C}$, $PT$)}\/} 
\newcommand{\Rules}{$\cal R$\hs}
\newcommand{\Table}{$\cal T$\/}
\newcommand{\CFG}{(\VN,~\VT,~$S$,~\Rules)\/} 
\newcommand{\ACFG}{(\VN,~\VT,~$S$,~\Rules,~$ATT$,~$M$,~$Q$)\/} 
\newcommand{\RBTSGw}{(\VN,~\VT,~$S$,~\Rules,~\A,~Viable?)\/} 
\newcommand{\SET}[2]{\mbox{\{#1~$|$~#2\}}\/}
\newcommand{\FUN}[3]{#1:~#2~\lra~#3\/}
\newcommand{\Root}[1]{$S_{#1}$\/}
\newcommand{\Rule}[2]{\mbox{#1~\lra~#2}\/}
\newcommand{\RuleEx}[2]{\mbox{$#1$~\lra~$#2$}\/}
\newcommand{\ROne}{\Rule{$N$}{$\alpha$}\/} 
\newcommand{\lmdir}{$\stackrel{*}{{\rightarrow}_{lm}}$}
\newcommand{\lmpdir}{$\stackrel{+}{{\rightarrow}_{lm}}$}
\newcommand{\DerOne}{\mbox{\Rule{\Root{t}}{$\alpha$}\lmdir$\beta$}} 
\newcommand{\VNTag}{\VN$^{'}$}
\newcommand{\VTTag}{\VT$^{'}$}
\newcommand{\STag}{$S^{'}$}
\newcommand{\RTag}{$R^{'}$}
\newcommand{\GTA}{(\VNTag,~\VTTag,~\STag,~\RTag)}
\newcommand{\AG}{\overline{A}}
\newcommand{\bIf}{{\bf If}~}
\newcommand{\bNot}{{\bf not}~}
\newcommand{\bThen}{{\bf Then}~}
\newcommand{\bElse}{{\bf Else}~}
\newcommand{\bFor}{{\bf For}~}
\newcommand{\bOf}{{\bf Of}~}
\newcommand{\bCase}{{\bf Case}~}
\newcommand{\bDo}{{\bf Do}~}
\newcommand{\bEnd}{{\bf End}~}
\newcommand{\bWhere}{{\bf Where}~}
\newcommand{\AND}{{\bf and}}
\newcommand{\OR}{{\bf or}~}
\newcommand{\bOtherwise}{{\bf Otherwise}~}
\newcommand{\mul}{\times}
\newcommand{\multi}{\times}
\newcommand{\parag}[1]{\paragraph{{\underline{#1}}}}
\newcommand{\sect}[1]{\section{#1}}
\newcommand{\Definition}[2]{\begin{description}\item[#1] {\it #2}\end{description}}
\newcommand{\DefinitionL}[2]{{#2}}
\newcommand{\Example}[1]{\paragraph{{\bf Example:}} {\it #1}}
%
\begin{abstract}
This paper presents a new view of Explanation-Based Learning (EBL) of 
natural language parsing. Rather than employing EBL for specializing parsers by 
inferring new ones, this paper suggests employing EBL for learning how to reduce ambiguity 
only partially.
We exemplify this by presenting a new EBL method that learns parsers that avoid 
spurious overgeneration, and we show how the same method can be used for reducing the 
sizes of stochastic grammars learned from tree-banks, e.g.~\cite{RENSDES,Charniak,Sekine}.

The present method consists of an EBL algorithm 
for learning partial-parsers, and a parsing algorithm which combines partial-parsers
with existing ``full-parsers". The learned partial-parsers, implementable as 
Cascades of Finite State Transducers (CFSTs), recognize and combine constituents efficiently,
prohibiting spurious overgeneration. The parsing algorithm combines a learned 
partial-parser with a given full-parser such that the role of the full-parser 
is limited to combining the constituents, recognized by the partial-parser, 
and to recognizing unrecognized portions of the input sentence. 
Besides the reduction of the parse-space prior to disambiguation, the present
method provides a way for refining existing disambiguation models that learn 
stochastic grammars from tree-banks e.g.~\cite{RENSDES,Charniak,Sekine}.

We exhibit encouraging empirical results using a pilot implementation:
parse-space is reduced substantially with minimal loss of coverage. The speedup 
gain for disambiguation models is exemplified by experiments with the DOP 
model~\cite{RENSDES}. 
\end{abstract}
%
\section{Introduction}
   Current work on natural language parsing is in large part directed towards 
   eliminating overgeneration of grammars by employing stochastic models for 
   disambiguation (e.g.~\cite{RENSDES,Sekine,Charniak}). 
   For many applications (e.g. Speech Understanding), probabilistic evaluation of the full parse-space using
   such models is NP-hard~\cite{Simaan96}, and even when it is deterministic polynomial-time,
   then grammar size is prohibitive. Therefore, it is necessary to develop methods that,
   on the one hand, reduce the space of analyses, as much as possible prior to disambiguation, 
   and on the other hand, reduce the sizes of grammars used for disambiguation. 
   This paper presents a method aimed at these two forms of reduction of time and space costs. 

   In recent work on speeding up parsing, effort is directed towards 
   {\em specializing}\/ broad-coverage grammar by~EBL 
   (e.g. \cite{Rayner88,Samuelsson94,RaynerCarter,SrinivasJoshi}).
   Grammar-specialization, in these works, amounts to replacing a given parser by a fresh
   efficient parser learned from the tree-bank. The learned parser trades coverage for
   efficiency. Inspired by these works, we present a new method based on EBL for learning 
   efficient parsers. Rather than specializing a given full-parser by inferring a new one, 
   the present method learns a partial-parser and combines it with the full-parser
   in a way that reduces ambiguity. The combination is a serial construction in which the 
   partial-parser is employed first for recognizing and combining constituents. 
   The partial-parser is learned such that it parses only those portions of the sentence that
   are ``safe" to parse, i.e. at the points where there is clear bias in the tree-bank. 
   These constituents are then passed through, together with unrecognized portions of the input, 
   to the full-parser, that completes the space only where necessary. 

   For disambiguation models such as~\cite{RENSDES,Sekine,Charniak}, the present method
   refines the cutting criteria which these models employ for inferring stochastic grammars. 
   This refinement results in the inference of smaller, yet no less powerful, statistical grammars.


\label{Def}
\section{Terminology}
\DefinitionL{A CFG derivation}
{A Context-Free Grammar (CFG) derivation is a sequence of one or more rewriting steps, 
starting with the start non-terminal of the grammar, employing the grammar productions. 
}
\DefinitionL{A subderivation}{A {\em subderivation}\/ is a subsequence of a derivation.}
\DefinitionL{Sentential-Form}
{A string of symbols which results from a CFG-derivation (of zero 
 or more rewriting steps) is called a {\em sentential-form}.
}
\DefinitionL{Partial-tree/subtree}
{A {\em partial-tree (also subtree)}\/ of a given tree $t$ is a tree-structure which 
 is the result of a subderivation of a derivation represented by $t$.}
\DefinitionL{Sentential partial-tree}
{A partial-tree which has as its root the start non-terminal is called a {\em sentential partial-tree}.}
\DefinitionL{Frontier of a partial-tree}
{The string obtained from the ordered sequence of leaf nodes of a partial-tree is 
called the {\em frontier}\/ of the partial-tree; the leaf nodes are called the
{\em frontier nodes}\/ of the partial-tree.
}
\DefinitionL{Appearance of CF-rule in a tree:}
{A~Context-Free~rule~\mbox{(CF-rule)}
$R$=\Rule{$A$}{$A_{1}\ldots A_{n}$}
is said to {\em appear}\/ in a tree $t$ if there is a node in $t$, labeled $A$, and 
that node has $n$ children labeled with (maintaining order from left to right) \mbox{$A_{1}\cdots A_{n}$}.
}
\DefinitionL{CFG underlying a tree-bank}
{The CFG \CFG~is called the {\em CFG underlying
a given tree-bank}\/ iff~  \Rules is the set
\(\SET{$R$}{rule $R$ appears in a tree in the tree-bank}\) (and the start non-terminal $S$,
non-terminal set $V_{N}$, terminal set $V_{T}$ are exactly those of the tree-bank).
A parser based on the CFG-underlying a tree-bank is called the Tree-bank parser (denoted
{\bf T-parser}).
}
\section{Explanation-Based Learning}
EBL~\cite{Mitchel,DeJong+Mooney,Harmelen+Bundy} 
is the name of a unifying 
framework for methods that learn from previously explained examples of 
a certain concept. EBL assumes a domain theory (or background theory) which provides explanations to
and enables the definition of concepts. In existing literature, the main goal of EBL 
is much faster recognition of concepts than the domain-theory does; EBL learns ``shortcuts" 
in computation (called macro-operators or ``chunks"), or   directives for changing 
the thread of computation. EBL stores the learned chunks in the form of partial-explanations 
to previously seen input instances, in order to apply them in the future 
to ``similar" input instances (in EBL, also ``similarity" is assumed provided by the domain-theory). 

The specification of  EBL consists of four preconditions and one postcondition.
The preconditions are: {\sl
1)~{\sl A domain theory:}
                        A description language for the domain at hand together with
                        rules and facts about the domain. 
2)~{\sl A target concept:}
                        A formal description, over the alfabet of the domain-theory,
                        of the to-be-learned relation. 
3)~{\sl An Operationality criterion:}
                       A requirement on the form of the target concept.
And~4)~{\sl training examples:} 
			A history which makes explicit the explanations
                        given by the domain-theory to examples that occurred in the past; 
                        the explanations consist of instances of the target concept. 
}
The postcondition is: {\sl
Find
  a generalization of the instances of the target concept given in the training-examples 
  that satisfies the operationality criterion. 
}

Past experience in Machine Learning cast doubts on the feasibility of improving performance 
by using EBL~\cite{Minton}.
Minton explains that EBL does not guarantee better performance, 
since the cost of applying the learned knowledge might outweigh the gain. 
Minton discusses a formula for computing the utility of knowledge during learning.
Generally speaking, this formulae is neither part of EBL nor part of the domain-theory; 
it is an extension to the EBL scheme by e.g. statistical inference over large sets of 
training examples. 
%
\section{Learning partial-parsers}
\label{Meth}
%
We assume a tree-bank representing a certain domain of application.
The tree-bank forms the training-examples of our EBL-based method, and
the linguistic annotation employed for annotating the sentences represents 
the domain-theory.
%
For the sake of presentation we delay the discussion of detail of the algorithm and
concentrate on a simplified version of it. The simplest instances of the target-concept of
our algorithm are called {\em probably always subsentential-forms}\/ (PA-SSFs).
\Definition{Subsentential-form}{ A subsentential-form (SSF) is a sequence of grammar-symbols
  which forms the frontier of a partial-tree.} 
\Definition{Probably always SSF}{An SSF \mbox{$ssf = N_{1}\cdots N_{m}$} 
  is called {\em Probably Always SSF (PA-SSF) with respect to the tree-bank}\/ 
  if the frequency of occurrence of \mbox{$N_{1}\cdots N_{m}$}
  in the tree-bank as SSF (denoted \mbox{$fc(N_{1}\cdots N_{m})$}) 
  is equal to the total frequency of its occurrence in the tree-bank 
  (denoted \mbox{$f(N_{1}\cdots N_{m})$}).
}

The concept PA-SSF formalizes the intuitive concept ``probably always constituent".
In reality, as discussed below, this concept is refined to become context-sensitive
and less rigid; it becomes ``probably {\em almost}\/ always constituent {\em in some local-context}".
Moreover, to avoid sparse-data problems we exclude the words of the language from the SSFs 
which we consider; the SSFs may consist of both part-of-speech tags (PoSTags) as well 
as phrase-symbols. 

\Definition{Associated subtree}{A partial-tree which has the sequence {\em ssf} as its frontier 
  is called a {\em subtree associated } with {\em ssf}. The set of subtrees associated with 
  {\em ssf}, with respect to a tree-bank, consists of all partial-trees of the tree-bank trees, 
  which are subtrees associated with {\em ssf}.
}

\subsection{The learning algorithm}
The goal of the algorithm is to learn the set of PA-SSFs
that represents the tree-bank trees in the {\em fastest and least ambiguous}\/ way possible.
The predicate ``least ambiguous" is instantiated in two ways:
1)~the learned (almost) PA-SSFs imply brackets which are most probably useful.
And~2)~the set of subtrees associated with a learned PA-SSF is assumed complete, i.e.
no more structures are necessary for future sentences containing that PA-SSF.
The second goal ``fastest" is implemented by selecting the 
PA-SSFs that reduce the tree-bank trees in the fastest way.
To achieve this we employ an operationality criterion which measures the utility of 
a PA-SSF. 
A measure of how much a single PA-SSF contributes to reducing a 
sentential-form is the {\em Reduction Factor}, and the ``expected utility" of a 
PA-SSF is estimated as the {\em Global Reduction Factor}: 

\Definition{Reduction Factor}{The {\em Reduction Factor}\/ ($RF$) of a given SSF {\em ssf}
is \mbox{$RF(ssf) = L(ssf) - 1$}, where $L(ssf)$ 
is the number of symbols which constitute {\em ssf}.
}

%
\Definition{Global RF}{The {\em global reduction factor}\/ of a given PA-SSF 
{\em ssf} with respect to the tree-bank is defined as 
\mbox{$GRF(ssf) = fc(ssf)\mul RF(ssf)$}, where $fc(ssf)$ is the frequency of {\em ssf}\/
as a constituent.
In case {\em ssf} is an SSF that is {\em not}\/ a PA-SSF  then {\em GRF(ssf)}~=~$-\infty$.
}

The specification of the learning algorithm is in figure~\ref{LearnAlg}. 
The algorithm learns PA-SSFs by an iterative procedure which ``eats" up the 
tree-bank trees from their leaves upwards.  Beginning with the tree-bank at hand, after 
each iteration, the procedure outputs: the set of learned PA-SSFs and a new tree-bank  
obtained by reducing all subtrees associated with a learned PA-SSF in all 
trees of the tree-bank at hand. In the next iteration, the same procedure is applied to the 
tree-bank output by this iteration. 
The procedure stops when there is nothing to learn anymore i.e. either there 
are no PA-SSFs to learn, or all tree-bank trees are fully reduced to their roots.


\Definition{Competitor SSF}{
Let $N$ be a node in tree $t$ and let $ssf$ be the partial-tree with $N$ as root.
The frontier of a partial-tree with root node which is a descendent or ancestor of 
$N$ in $t$ is called a {\em competitor} of $ssf$.
}
\Definition{Operationality criterion}{
At each iteration of the algorithm, for each sentential partial-tree $t$ in the tree-bank,
for each SSF {\em ssf}\/ in $t$, {\em ssf} is learned iff {\em ssf}\/ is PA-SSF and 
\mbox{$GRF(ssf) \geq GRF(x)$}, for all $x$ which is a competitor of {\em ssf} in $t$.
}

Consider again the specification in figure~\ref{LearnAlg}. 
Let the algorithm be at a certain iteration $i$
and let each node in each partial-tree of the current tree-bank (${\TB}_{i}$) have a unique address. 
Also define the global reduction factor of an address $N$ of a node, $GRF(N)$, to be equal to 
{\em GRF(ssf)}, where ssf is the SSF on the frontier of the partial-tree under the node with 
address $N$. The operationality criterion is implemented in the specification at step (2.), where
nodes are marked. The learned PA-SSFs are those SSFs which form the frontiers of partial-trees
of which the root is a node which was marked at some iteration.
\newcommand{\COMSIZE}{}
\newcommand{\comment}[1]{{\COMSIZE\tt\bf ~/*~#1~~*/}}
\newcommand{\bcomment}[1]{{\COMSIZE\tt\bf ~/*~#1}}
\newcommand{\ecomment}[1]{{\COMSIZE\tt\bf ~#1~~*/}}
\newcommand{\EMPTY}{$\emptyset$}
\newcommand{\InBf}[1]{{\bf #1}}
\newcommand{\FUNC}{{\bf Function}~}
\newcommand{\BEGIN}{{\bf b}\={\bf egin}~}
\newcommand{\END}{{\bf end}~}
\newcommand{\UNION}{{\bf ${\bf \cup}$}~}
\newcommand{\REPEAT}{{\bf r}\={\bf epeat}~}
\newcommand{\UNTIL}{{\bf until}~}
\newcommand{\Set}{{\bf Set}~}
\newcommand{\TreeBank}{{\bf TreeBank}~}
\newcommand{\Integer}{{\bf Integer}~}
\newcommand{\bReturn}{{\bf return}~}
\newcommand{\bReturns}{{\bf returns}~}
\newcommand{\Boolean}{{\bf Boolean}~}
\newcommand{\For}{{\bf F}\={\bf or}~}

\begin{figure*}[thb]
\epsfxsize=14cm
\epsfbox{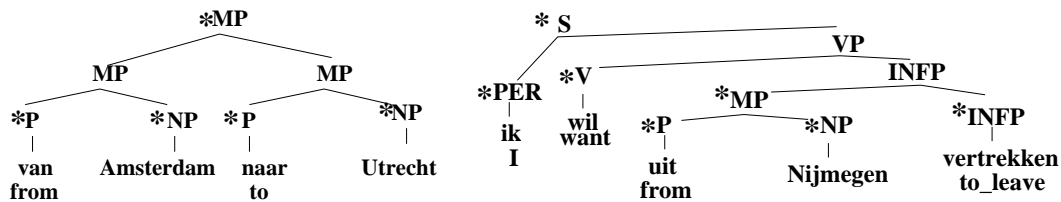}
\caption{Two trees marked by the learning algorithm}
\label{uone}
\end{figure*}
\newcommand{\bREPEAT}{{\bf Repeat}}
\newcommand{\bUNTIL}{{\bf until}}
\begin{figure*}[thb]
\hrule
{\sl
\begin{tabbing}
\comment{Let $N$ denote a unique address of a node of a tree $t$. Also let ${\TB}_{i}$ denote the}\\
\comment{tree-bank obtained after $i$ iterations, where ${\TB}_{0}$ denotes the given tree-bank.}\\ 
\comment{{\bf Frontier\_Of($N$)} denotes the frontier (i.e. an SSF) of the partial-tree under $N$.}\\
\comment{{\bf Descendent(Nch,Np)} denotes the predicate: the node addressed $Nch$ is a ~~~~~}\\
\comment{descendent of the node addressed $Np$.~~~~~~~~~~~~~~~~~~~~~~~~~~~~~~~~~~~~~~~~~~~~~~~~~~~}
\end{tabbing}
\begin{description}
\item [1.] $i := 0;$
\item [\bREPEAT]
\item [2.] 
\begin{tabbing}
$\forall$ \= $t\in {\TB}_{i}$, ~~$\forall$ node address $N$ in $t$:  N is {\it marked}\/ {\bf iff} \\
          \> Frontier\_Of(N) is PA-SSF in ${\TB}_{i}$~ ~\AND \\
          \> ~~~$\forall$\=$Nx\neq N$ in $t$: ( Descendent(Nx,N) \OR Descendent(N,Nx) )~\lra~ (GRF(N) $>$ GRF(Nx)); 
\end{tabbing}
\item [{\bf 3.}] $i := i+1$;
\item [{\bf 4.}] ${\TB}_{i}$ := (${\TB}_{i-1}$ after reducing all partial-trees under marked nodes); 
\item [\bUNTIL] ((${\TB}_{i} == \emptyset$)  \OR~ (${\TB}_{i} == {\TB}_{i-1}$));
\end{description}
}
\hrule
\caption{The Learning Algorithm}
\label{LearnAlg}
\end{figure*}
%
\subsubsection{Detail of learning algorithm}
Now we present further detail of the algorithm.
The term ``PA-SSF" is redefined as follows: 
\begin{verse}
{\em ssf} is called PA-SSF if it fulfills \mbox{$\frac{fc(ssf)}{f(ssf)} \geq \theta$},\\
where ~\mbox{$0 < \theta \leq 1$} is a threshold.
\end{verse}
This definition of PA-SSF makes the target-concept of our
EBL method become ``with probability more than $\theta$ a constituent".
The algorithm employs this definition as follows. A threshold is set on the values of $\theta$, where 
$\theta$ is allowed to change during learning (the default value of this threshold is $\theta~=~1.0$
unless stated otherwise). Suppose the threshold on $\theta$ is 
0.75. The algorithm starts in the first iteration with learning PA-SSFs of 
\mbox{$\theta = 1.0$}. Each time there are no more PA-SSFs to learn, under the current value
of $\theta$, it reduces $\theta$ by a fixed amount (e.g. 0.05) until $\theta$ becomes equal to the 
threshold (0.75). Then the algorithm stops learning.

We also employ a threshold ($\tau$) on the minimum frequency of SSFs; 
an SSF must be frequent enough in order to qualify for the PA-SSF test. 
Currently this threshold is set at the maximum of a fixed integer (e.g. 10)
and a percentage of the number of trees in the tree-bank (e.g. 0.3\%). However, a more
principled way to set the threshold is by letting it be a function of
the distribution of SSFs in the tree-bank.

The algorithm also employs a definition of PA-SSF conditioned on
local context, rather than fully context-free:
\begin{verse}
A sequence of symbols is called PA-SSF in context C iff the ratio 
between its frequency as SSF in context~C and its total frequency in context~C is~$\geq~\theta$.
\end{verse}
The local context that is employed consists of four fields: two grammar symbols to 
the left of and two to the right of an SSF. Since after the first round of the 
learning algorithm the training material consists of sentential partial-trees, this 
kind of local context may consist of PoSTags as well as {\em phrasal symbols}.
The algorithm can use this local context in order to enhance learning and parsing.
In the current implementation, however, we employ this local context only during learning and
in a quite simplistic manner. 

Since currently local-context is not employed during parsing, the learning algorithm 
is tuned to prefer as general local-contexts as possible.
The learning algorithm assumes in the first place that all four 
fields of the local-context of an SSF are wild-cards. In case the SSF is not a PA-SSF 
in that context, then the algorithm {\em retreats}\/ and assumes any three of the four 
fields to be wild-cards. In case an SSF is not a PA-SSF under three or more wild-cards of
local-context then it is not learned, i.e.  two or less wild-card local-contexts 
do not contribute to learning.
Future implementations, however, shall have to take this local-context more seriously 
both in learning and in parsing.
\Example
{In figure~\ref{uone}, two example trees are shown. The asterisks
 in the figure denote the borders of subtrees associated with PA-SSFs
 learned from the tree-bank. The sequences of symbols marked with an asterisk
 at the frontier of a subtree, which has a marked root, form the learned PA-SSFs.
 In the tree at the left-hand side of figure\ref{uone}, there is only one
 PA-SSF that reduces the tree: \mbox{($p$ $np$ $p$ $np$)}, which corresponds
 to ``from~Amsterdam~to~Utrecht". In the right-hand side tree of figure~\ref{uone},
 there are two PA-SSFs, ($p~np$) and ($per~v~mp~infp$). 

  In the first iteration, the learning algorithm reduced the left tree totally and reduced 
  the right tree only at the constituent ``from Nijmegen". In the second 
  iteration, the leftovers of the right-tree, a sentential partial-tree with frontier
  ``ik/I wil/want mp vertrekken/to\_leave", is reduced fully. 
  If there are other partial-trees which are left over in the tree-bank, after these two iterations, 
  then the algorithm will attempt reducing them in subsequent iterations.
  If there are no more PA-SSFs to learn, the algorithm stops (possibly leaving
  some partial-trees not fully reduced).
}
\subsection{The parsing algorithm}
\label{ParAlg}
A Tree-Substitution Grammar (TSG) is a CFG with rules which are partial-trees called
elementary-trees.  Let the set of subtrees associated with the PA-SSFs, which the
learning algorithm outputs, be the set of elementary-trees of a TSG; 
the TSG has the same start-symbol, terminal and non-terminal symbols as the CFG underlying 
the tree-bank. This TSG is employed as a partial-parser (other implementations are 
discussed below). 

The new parsing algorithm combines the partial-parser with a given full-parser.
It has two stages: firstly it employs the partial-parser for parsing 
the input sentence bottom-up, resulting in a space of partial-parses  
combined from subtrees associated with PA-SSFs. In the second phase it
employs a given full-parser to complete these partial parses into full parses. 
Crucially, the second phase of the algorithm takes advantage of the construction of
the learning algorithm. It makes two assumptions concerning the space of partial parses 
which the partial-parser constructed:
\begin{itemize}
\item If a sequence of symbols is recognized by the partial-parser then it is 
  highly probable that all its subtrees are present in the chart 
  (as these are either associated subtrees or combinations of associated subtrees). 
  Thus it is not necessary to attempt reparsing portions of the sentence which 
  were recognized as by the partial-parser.
\item
  In the default case, \mbox{$\theta = 1.0$}, a PA-SSF implies 
  ``sure" constituent-borders; therefore, brackets placed by the full-parser are 
  not allowed to cross the borders of a PA-SSF. In case two PA-SSFs cross each other,
  a highly unlikely case, then both PA-SSFs are removed from the partial-parser's output. 
\end{itemize}
Thus, the task of the full-parser is limited to parsing
{\em totally uncovered portions and combining them with the partial-trees 
provided by the partial-parser in ways that do not cross recognized PA-SSFs
with \mbox{$\theta = 1.0$}}. 
In this paper we employ the CFG underlying the tree-bank (i.e. T-parser) as the full-parser.
%
%
\subsubsection{Implementation of parsing algorithm}
\label{ImpPAlg}
The current pilot implementation of the partial-parser does not take local 
context of PA-SSFs into consideration. The partial-parser is implemented 
as a parser for TSGs~\cite{Simaan95}, based on an extension to the CYK algorithm~\cite{CKY}).
However, the partial-parser can be implemented as a Cascade of 
Finite State Transducers (CFSTs). 
A Finite State Transducer (FST) is learned at each iteration of the learning algorithm; the FST's
language is the set of PA-SSFs learned at that iteration, and the output of
the FST on recognition of a PA-SSF is the set of subtrees associated with that
PA-SSF.
%
\section{Existing related methods}
\label{Oth}
EBL~was introduced to NLP by Rayner~\cite{Rayner88}; Rayner
employs EBL for specializing broad-coverage grammars to specific domains. 
In \cite{RaynerSamuelsson,RaynerCarter} grammar specialization is conducted by
chunking the trees of a tree-bank according to ``chunking criteria"
which are manually specified e.g. chunks correspond to trees with
roots which correspond to full utterances, NPs, PPs or non-recursive NPs.
Samuelsson~\cite{Samuelsson94} is the first to depart from manual 
specification of chunking criteria in NLP; the chunking of the tree-bank 
trees employs the information theoretic measure of entropy.
Samuelsson measures the entropy of a grammar non-terminal as the measure 
of how hard it is to decide on the choice of the next rule application 
given that non-terminal. Then he marks the nodes with the largest entropy 
as cutting nodes using an iterative algorithm. 
In~\cite{SrinivasJoshi} the specific structure of the Lexicalized 
Tree-Adjoining Grammar (LTAG) derivations is exploited to result in an 
EBL method specific for LTAG. This differs from the other efforts in that 
the generalization which they employ is not limited only to 
{\em goal-regression}\/ but allows generalizing the structure of explanations. 
Their method learns from the LTAG derivations of the training-examples all 
sequences of PoSTags and reduces those to regular-expressions by generalizing 
on sequences of adjunctions with a Kleene-star; the generalized LTAG-derivations 
are stored indexed by the PoSTag sequences.

{\sl Relation to Samuelsson's EBL:}~
The present method is similar to Samuelsson's in that it learns 
``cutting criteria" from the data. Our method differs from Samuelsson's
in that the cutting criteria are computed from an opposite direction.
Samuelsson's maximum entropy is aimed at maximizing coverage,
and his approach is derivational since the entropy is computed 
on steps of derivations starting from the start non-terminal. 
The target concept of our method is a PA-SSF not a non-terminal
(i.e. ``probably always constituent" vs. ``constituent" resp.).
Our method assumes a reductive approach and results in a partial-parser rather 
than a specialized parser. 

{\sl Relation to LTAG's EBL:}~
The concept of PA-SSF employed by our method is a generalization of 
the sentential PoSTag sequences employed in~\cite{SrinivasJoshi}. Our
method can be easily extended to accommodate LTAG generalizations
of derivations and of PA-SSFs; to this end it is necessary to have a tree-bank 
annotated with LTAG derivations. The subtrees associated with learned 
PA-SSFs are then generalized partial derivations of LTAG.
\section{Application to DOP}
\label{DOPasEBL}
This section relates the present EBL method to existing models of disambiguation 
that project stochastic grammars from tree-banks, e.g.~\cite{RENSDES,Charniak,Sekine}. 
To this end, we firstly relate these models to EBL, and then
show that our new EBL method refines these models.

We are concerned only with models that project the same 
grammatical description as that employed for annotation of the tree-bank. Among these
models, the Data Oriented Parsing (DOP) model~\cite{Scha,RENSDES} takes the most 
radical point of view. DOP projects all partial-trees from a tree-bank and employs them 
as a stochastic grammar called a Stochastic Tree-Substitution Grammar (STSG).
Other models in the same category are presented in~\cite{Charniak,Sekine}. 
Charniak~\cite{Charniak} employs the tree-bank for projecting 
Stochastic CFGs (SCFGs).  And~\cite{Sekine} presents a
constrained DOP-like model which projects STSGs; cutting the tree-bank
trees takes place only at nodes labeled either with $S$ or with $NP$.
In this section we concentrate on DOP since it constitutes a generalization
of the other two efforts.

In~\cite{RENSDES}, the specification of DOP is as follows. A DOP model
has four parameters: 
\begin{enumerate}
\item ~sentence-analyses, i.e. syntactically labeled phrase structure
trees given in a tree-bank, 
\item sub-analyses, i.e. partial-trees, 
\item combination-operations, i.e. substitution, and
\item combination-probabilities.
\end{enumerate}
 The rest of the definition of the DOP model
concerns how to infer probabilities of partial-trees from the tree-bank, and how to compute 
probabilities of combinations of partial-trees. The instantiation of DOP as realized 
in~\cite{RENSDES} is an STSG, which has the set of {\em all}\/ partial-trees
of the tree-bank trees as elementary-trees. We shall not give further details of DOP 
since this is out of the scope of this paper. 

Let us rewrite Bod's specification using the terminology of EBL. Firstly,
the so called domain-theory consists of the annotation convention as well as the 
annotation intuitions used for the annotation of the tree-bank.
The tree-bank contains sentences and their tree structures: the trees
constitute ``explanations" (proofs) given by the domain-theory to the 
fact that the sequences of words on their frontiers are sentences.
The target-concept of DOP is the concept of a 
constituent, represented by non-terminals of the tree-bank trees. 
The sub-analysis used by DOP are simply partial-trees, which form instances of the target-concept. 
These partial-trees are obtained by using a simple operationality criterion, which states that 
any partial-tree obtained from a tree-bank tree is acceptable 
(in the experiments mentioned in~\cite{RENSDES}, Bod limits the depth of partial-trees,
 Charniak~\cite{Charniak} limits the partial-trees to CFG rules, and in~\cite{Sekine} only 
 a subset of the non-terminals are allowed to supply partial-trees). 
The combination-operation of DOP is inherent to the 
assumption that the theory (phrase structure grammar) employs that operation.
The fourth parameter of DOP, i.e. the inference and 
the definitions of probabilities of combinations of partial-trees, extends the EBL scheme.
This extension enables DOP, and the other models mentioned above, to apply statistical analysis 
over large sets of trees in order to facilitate disambiguation. 
The interesting part of viewing these models in EBL terminology is the fact that these 
models do not aim at speedup, but rather at the memory-based behavior of EBL. 

The new EBL method can be used in order to define the operationality criterion
for DOP as follows. 
\begin{itemize}
\item Apply the algorithm in figure~\ref{LearnAlg} to the given tree-bank. The result is
     the same tree-bank except that now there are marking on nodes which delimit the
     subtrees associated with the learned PA-SSFs. 
\item Mark also all nodes which are {\em not}\/ internal to any subtree associated with 
      a learned PA-SSF. And mark all PosTag nodes in all tree-bank trees. 
\end{itemize}
If learning was successful, then only some of the nodes of the tree-bank trees are marked now.
The operationality criterion for DOP is then: 
\begin{quote}
{\it A partial-tree is projected iff its 
root and the nodes on its frontier are marked, i.e. cutting the trees for DOP is not allowed 
at unmarked nodes}. Crucially, this way of cutting allows the projection of 
partial-trees which are {\em combinations}\/ of subtrees associated with learned PA-SSFs.
\end{quote}

The main remaining question on this refinement of DOP concerns the probabilities of the 
partial-trees projected from the tree-bank. In DOP, the probability of a partial-tree with
a root labeled $N$ is defined as the ratio between its frequency and the total frequency of 
all partial-trees that have $N$ as their root-label. Since the space of partial-trees is
smaller in the refinement, the probabilities will be different than in the original DOP. 
We conjecture that due to reducing the number of parameters of the model, sparse-data
effects should be reduced (future work shall address this issue).
\section{Empirical results}
\label{Exp}
%
The present method was developed within a Dutch national project on
a {\em dialogue system concerning public-transportation 
information}\/ (called OVIS) (http://grid.let.rug.nl:4321/). 
Within the project, a vast amount of dialogues were collected,
and the user's utterances were syntactically and semantically annotated~\cite{DOPNWORep}.
For experimentation we employ a tree-bank of the first 5000 syntactically 
annotated utterances. 
Here we only report experiments on parsing transcribed 
utterances\footnote{The present method was applied together with DOP for parsing
word-graphs in a speech recognition-task and resulted in, compared to DOP, on average 
speedup of 10 times with virtually no loss of accuracy. 
Average speedup for word-graphs containing more than 40 states exceeds 20 times.}.
\begin{figure}[ht]
\epsfxsize=7.5cm
\epsfbox{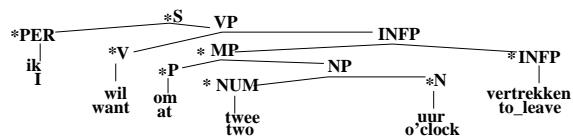}
\caption{Another tree from OVIS}
\label{ucombone}
\end{figure}

The annotation of the OVIS tree-bank is exemplified by the trees in figures~\ref{uone} and \ref{ucombone}.
Due to the fact that OVIS contains answers to questions within a dialogue system, the
sentences are often short but surprisingly variable in structure;  many of these sentences 
contain repetitions, corrections and strange constructions (usually rendered ungrammatical
by linguistic theories). 
Below we report on two sets of experiments.  The first set observes the learning curves
of the present EBL method by combining the learned partial-parsers with a 
T-parser (i.e. CFG). And the second set studies the refinement of DOP using the present EBL method.
All timing experiments were conducted on SGI Indigo with 640~MB RAM.
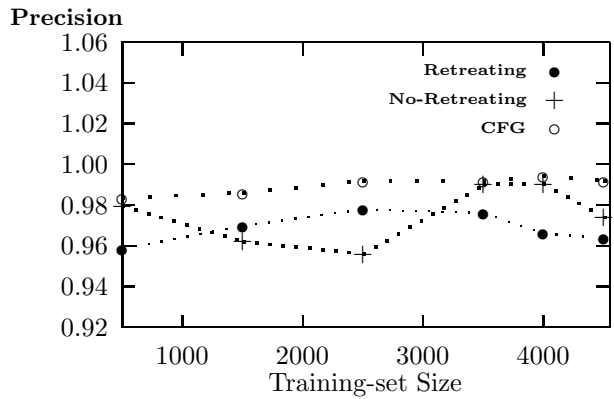
\begin{figure}[tbh]
\setlength{\unitlength}{0.240900pt}
\ifx\plotpoint\undefined\newsavebox{\plotpoint}\fi
\sbox{\plotpoint}{\rule[-0.200pt]{0.400pt}{0.400pt}}%
\begin{picture}(1049,584)(0,0)
\font\gnuplot=cmr10 at 10pt
\gnuplot
\sbox{\plotpoint}{\rule[-0.200pt]{0.400pt}{0.400pt}}%
\put(220.0,113.0){\rule[-0.200pt]{4.818pt}{0.400pt}}
\put(198,113){\makebox(0,0)[r]{$0.92$}}
\put(965.0,113.0){\rule[-0.200pt]{4.818pt}{0.400pt}}
\put(220.0,177.0){\rule[-0.200pt]{4.818pt}{0.400pt}}
\put(198,177){\makebox(0,0)[r]{$0.94$}}
\put(965.0,177.0){\rule[-0.200pt]{4.818pt}{0.400pt}}
\put(220.0,241.0){\rule[-0.200pt]{4.818pt}{0.400pt}}
\put(198,241){\makebox(0,0)[r]{$0.96$}}
\put(965.0,241.0){\rule[-0.200pt]{4.818pt}{0.400pt}}
\put(220.0,305.0){\rule[-0.200pt]{4.818pt}{0.400pt}}
\put(198,305){\makebox(0,0)[r]{$0.98$}}
\put(965.0,305.0){\rule[-0.200pt]{4.818pt}{0.400pt}}
\put(220.0,369.0){\rule[-0.200pt]{4.818pt}{0.400pt}}
\put(198,369){\makebox(0,0)[r]{$1.00$}}
\put(965.0,369.0){\rule[-0.200pt]{4.818pt}{0.400pt}}
\put(220.0,433.0){\rule[-0.200pt]{4.818pt}{0.400pt}}
\put(198,433){\makebox(0,0)[r]{$1.02$}}
\put(965.0,433.0){\rule[-0.200pt]{4.818pt}{0.400pt}}
\put(220.0,497.0){\rule[-0.200pt]{4.818pt}{0.400pt}}
\put(198,497){\makebox(0,0)[r]{$1.04$}}
\put(965.0,497.0){\rule[-0.200pt]{4.818pt}{0.400pt}}
\put(220.0,561.0){\rule[-0.200pt]{4.818pt}{0.400pt}}
\put(198,561){\makebox(0,0)[r]{$1.06$}}
\put(965.0,561.0){\rule[-0.200pt]{4.818pt}{0.400pt}}
\put(314.0,113.0){\rule[-0.200pt]{0.400pt}{4.818pt}}
\put(314,68){\makebox(0,0){$1000$}}
\put(314.0,541.0){\rule[-0.200pt]{0.400pt}{4.818pt}}
\put(503.0,113.0){\rule[-0.200pt]{0.400pt}{4.818pt}}
\put(503,68){\makebox(0,0){$2000$}}
\put(503.0,541.0){\rule[-0.200pt]{0.400pt}{4.818pt}}
\put(692.0,113.0){\rule[-0.200pt]{0.400pt}{4.818pt}}
\put(692,68){\makebox(0,0){$3000$}}
\put(692.0,541.0){\rule[-0.200pt]{0.400pt}{4.818pt}}
\put(881.0,113.0){\rule[-0.200pt]{0.400pt}{4.818pt}}
\put(881,68){\makebox(0,0){$4000$}}
\put(881.0,541.0){\rule[-0.200pt]{0.400pt}{4.818pt}}
\put(220.0,113.0){\rule[-0.200pt]{184.288pt}{0.400pt}}
\put(985.0,113.0){\rule[-0.200pt]{0.400pt}{107.923pt}}
\put(220.0,561.0){\rule[-0.200pt]{184.288pt}{0.400pt}}
\put(133,600){\makebox(0,0){{\bf\footnotesize Precision}}}
\put(602,23){\makebox(0,0){{Training-set Size}}}
\put(220.0,113.0){\rule[-0.200pt]{0.400pt}{107.923pt}}
\put(855,515){\makebox(0,0)[r]{{\tiny\bf Retreating}}}
\put(899,515){\raisebox{-.8pt}{\makebox(0,0){$\bullet$}}}
\put(220,236){\raisebox{-.8pt}{\makebox(0,0){$\bullet$}}}
\put(409,272){\raisebox{-.8pt}{\makebox(0,0){$\bullet$}}}
\put(598,298){\raisebox{-.8pt}{\makebox(0,0){$\bullet$}}}
\put(787,292){\raisebox{-.8pt}{\makebox(0,0){$\bullet$}}}
\put(881,260){\raisebox{-.8pt}{\makebox(0,0){$\bullet$}}}
\put(976,253){\raisebox{-.8pt}{\makebox(0,0){$\bullet$}}}
\put(220,236){\usebox{\plotpoint}}
\multiput(220,236)(20.389,3.884){10}{\usebox{\plotpoint}}
\multiput(409,272)(20.562,2.829){9}{\usebox{\plotpoint}}
\multiput(598,298)(20.745,-0.659){9}{\usebox{\plotpoint}}
\multiput(787,292)(19.648,-6.689){5}{\usebox{\plotpoint}}
\multiput(881,260)(20.699,-1.525){4}{\usebox{\plotpoint}}
\put(976,253){\usebox{\plotpoint}}
\sbox{\plotpoint}{\rule[-0.400pt]{0.800pt}{0.800pt}}%
\put(855,470){\makebox(0,0)[r]{{\tiny\bf No-Retreating}}}
\put(899,470){\makebox(0,0){$+$}}
\put(220,303){\makebox(0,0){$+$}}
\put(409,247){\makebox(0,0){$+$}}
\put(598,228){\makebox(0,0){$+$}}
\put(787,337){\makebox(0,0){$+$}}
\put(881,337){\makebox(0,0){$+$}}
\put(976,285){\makebox(0,0){$+$}}
\sbox{\plotpoint}{\rule[-0.500pt]{1.000pt}{1.000pt}}%
\put(220,303){\usebox{\plotpoint}}
\multiput(220,303)(19.900,-5.896){10}{\usebox{\plotpoint}}
\multiput(409,247)(20.651,-2.076){9}{\usebox{\plotpoint}}
\multiput(598,228)(17.980,10.369){11}{\usebox{\plotpoint}}
\multiput(787,337)(20.756,0.000){4}{\usebox{\plotpoint}}
\multiput(881,337)(18.206,-9.966){5}{\usebox{\plotpoint}}
\put(976,285){\usebox{\plotpoint}}
\sbox{\plotpoint}{\rule[-0.600pt]{1.200pt}{1.200pt}}%
\put(855,426){\makebox(0,0)[r]{{\tiny\bf CFG}}}
\put(899,426){\raisebox{-.8pt}{\makebox(0,0){$\circ$}}}
\put(220,316){\raisebox{-.8pt}{\makebox(0,0){$\circ$}}}
\put(409,324){\raisebox{-.8pt}{\makebox(0,0){$\circ$}}}
\put(598,343){\raisebox{-.8pt}{\makebox(0,0){$\circ$}}}
\put(787,343){\raisebox{-.8pt}{\makebox(0,0){$\circ$}}}
\put(881,350){\raisebox{-.8pt}{\makebox(0,0){$\circ$}}}
\put(976,343){\raisebox{-.8pt}{\makebox(0,0){$\circ$}}}
\sbox{\plotpoint}{\rule[-0.500pt]{1.000pt}{1.000pt}}%
\put(220,316){\usebox{\plotpoint}}
\multiput(220,316)(41.474,1.756){5}{\usebox{\plotpoint}}
\multiput(409,324)(41.303,4.152){5}{\usebox{\plotpoint}}
\multiput(598,343)(41.511,0.000){4}{\usebox{\plotpoint}}
\multiput(787,343)(41.396,3.083){2}{\usebox{\plotpoint}}
\multiput(881,350)(41.399,-3.050){3}{\usebox{\plotpoint}}
\put(976,343){\usebox{\plotpoint}}
\end{picture}
\caption{Learning curves for parser-precision} 
\label{FigPlot1}
\end{figure}
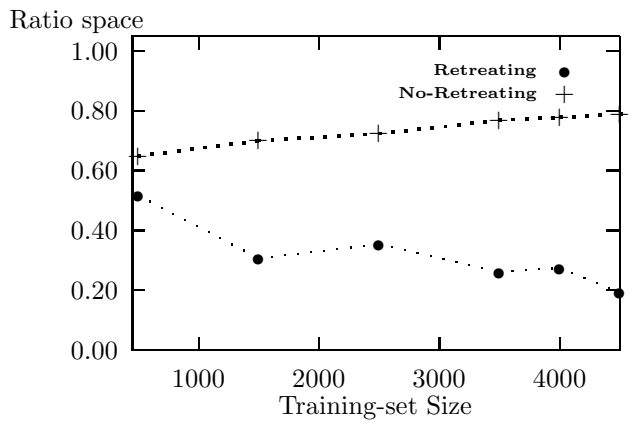
\begin{figure}[bth]
\setlength{\unitlength}{0.240900pt}
\ifx\plotpoint\undefined\newsavebox{\plotpoint}\fi
\sbox{\plotpoint}{\rule[-0.200pt]{0.400pt}{0.400pt}}%
\begin{picture}(1049,629)(0,0)
\font\gnuplot=cmr10 at 10pt
\gnuplot
\sbox{\plotpoint}{\rule[-0.200pt]{0.400pt}{0.400pt}}%
\put(220.0,113.0){\rule[-0.200pt]{184.288pt}{0.400pt}}
\put(220.0,113.0){\rule[-0.200pt]{4.818pt}{0.400pt}}
\put(198,113){\makebox(0,0)[r]{$0.00$}}
\put(965.0,113.0){\rule[-0.200pt]{4.818pt}{0.400pt}}
\put(220.0,207.0){\rule[-0.200pt]{4.818pt}{0.400pt}}
\put(198,207){\makebox(0,0)[r]{$0.20$}}
\put(965.0,207.0){\rule[-0.200pt]{4.818pt}{0.400pt}}
\put(220.0,301.0){\rule[-0.200pt]{4.818pt}{0.400pt}}
\put(198,301){\makebox(0,0)[r]{$0.40$}}
\put(965.0,301.0){\rule[-0.200pt]{4.818pt}{0.400pt}}
\put(220.0,395.0){\rule[-0.200pt]{4.818pt}{0.400pt}}
\put(198,395){\makebox(0,0)[r]{$0.60$}}
\put(965.0,395.0){\rule[-0.200pt]{4.818pt}{0.400pt}}
\put(220.0,489.0){\rule[-0.200pt]{4.818pt}{0.400pt}}
\put(198,489){\makebox(0,0)[r]{$0.80$}}
\put(965.0,489.0){\rule[-0.200pt]{4.818pt}{0.400pt}}
\put(220.0,583.0){\rule[-0.200pt]{4.818pt}{0.400pt}}
\put(198,583){\makebox(0,0)[r]{$1.00$}}
\put(965.0,583.0){\rule[-0.200pt]{4.818pt}{0.400pt}}
\put(324.0,113.0){\rule[-0.200pt]{0.400pt}{4.818pt}}
\put(324,68){\makebox(0,0){$1000$}}
\put(324.0,586.0){\rule[-0.200pt]{0.400pt}{4.818pt}}
\put(513.0,113.0){\rule[-0.200pt]{0.400pt}{4.818pt}}
\put(513,68){\makebox(0,0){$2000$}}
\put(513.0,586.0){\rule[-0.200pt]{0.400pt}{4.818pt}}
\put(702.0,113.0){\rule[-0.200pt]{0.400pt}{4.818pt}}
\put(702,68){\makebox(0,0){$3000$}}
\put(702.0,586.0){\rule[-0.200pt]{0.400pt}{4.818pt}}
\put(891.0,113.0){\rule[-0.200pt]{0.400pt}{4.818pt}}
\put(891,68){\makebox(0,0){$4000$}}
\put(891.0,586.0){\rule[-0.200pt]{0.400pt}{4.818pt}}
\put(220.0,113.0){\rule[-0.200pt]{184.288pt}{0.400pt}}
\put(985.0,113.0){\rule[-0.200pt]{0.400pt}{118.764pt}}
\put(220.0,606.0){\rule[-0.200pt]{184.288pt}{0.400pt}}
\put(133,629){\makebox(0,0){{Ratio space}}}
\put(602,23){\makebox(0,0){{Training-set Size}}}
\put(220.0,113.0){\rule[-0.200pt]{0.400pt}{118.764pt}}
\put(855,551){\makebox(0,0)[r]{{\tiny\bf Retreating}}}
\put(899,551){\raisebox{-.8pt}{\makebox(0,0){$\bullet$}}}
\put(229,356){\raisebox{-.8pt}{\makebox(0,0){$\bullet$}}}
\put(418,257){\raisebox{-.8pt}{\makebox(0,0){$\bullet$}}}
\put(607,279){\raisebox{-.8pt}{\makebox(0,0){$\bullet$}}}
\put(796,236){\raisebox{-.8pt}{\makebox(0,0){$\bullet$}}}
\put(891,242){\raisebox{-.8pt}{\makebox(0,0){$\bullet$}}}
\put(985,204){\raisebox{-.8pt}{\makebox(0,0){$\bullet$}}}
\put(229,356){\usebox{\plotpoint}}
\multiput(229,356)(18.386,-9.631){11}{\usebox{\plotpoint}}
\multiput(418,257)(20.616,2.400){9}{\usebox{\plotpoint}}
\multiput(607,279)(20.238,-4.604){9}{\usebox{\plotpoint}}
\multiput(796,236)(20.714,1.308){5}{\usebox{\plotpoint}}
\multiput(891,242)(19.243,-7.779){5}{\usebox{\plotpoint}}
\put(985,204){\usebox{\plotpoint}}
\sbox{\plotpoint}{\rule[-0.400pt]{0.800pt}{0.800pt}}%
\put(855,515){\makebox(0,0)[r]{{\tiny\bf No-Retreating}}}
\put(899,515){\makebox(0,0){$+$}}
\put(229,418){\makebox(0,0){$+$}}
\put(418,442){\makebox(0,0){$+$}}
\put(607,453){\makebox(0,0){$+$}}
\put(796,474){\makebox(0,0){$+$}}
\put(891,479){\makebox(0,0){$+$}}
\put(985,483){\makebox(0,0){$+$}}
\sbox{\plotpoint}{\rule[-0.500pt]{1.000pt}{1.000pt}}%
\put(229,418){\usebox{\plotpoint}}
\multiput(229,418)(20.590,2.615){10}{\usebox{\plotpoint}}
\multiput(418,442)(20.720,1.206){9}{\usebox{\plotpoint}}
\multiput(607,453)(20.629,2.292){9}{\usebox{\plotpoint}}
\multiput(796,474)(20.727,1.091){5}{\usebox{\plotpoint}}
\multiput(891,479)(20.737,0.882){4}{\usebox{\plotpoint}}
\put(985,483){\usebox{\plotpoint}}
\end{picture}
\caption{{\footnotesize Learning curves: active-nodes~$\frac{EBL+T\_Parser}{T\_Parser}$}} 
\label{FigPlot2}
\end{figure}
\subsection{OVIS experiments with T-parser}
The experiments concern both coverage as well as size of parse-space. 
We employ the T-parser underlying the tree-bank (CFG) as a full-parser.
In table~\ref{Tab1} we list the results of ten independent experiments,
each obtained by a random split of 4500 training-set and 500 test-set.
Since the domain contains many (easy for parsing) one word utterances 
(e.g. "yes" or "no"), we exclude one word utterances from the results.
On average, the ten test-sets contained 337.2 (of 500) utterances 
longer than one word. Table~\ref{Tab1} shows the results on utterances 
longer than one word, with mean length of 5.57 words per utterance.
For training the EBL learning algorithm we set a threshold on the frequency 
of SSFs: 0.3\% of the size of the training-set (i.e. 14). To avoid problems 
of unknown words, we allowed the words of the test-set to be included with 
{\em all}\/ postags with which they appear in the whole tree-bank (for
both parsers). 
\newcommand{\inchart}{{in chart}}
\begin{table*}[htb]
\begin{tabular}{|c|c|c|c|c|}
\hline
Parser                      & Right parse \inchart      & Any parse \inchart        & Precision & Active nodes \\
\hline
{\small\bf T\_parser}       & 97.78\% (1.1\%) & 99.62\% (0.3\%) & 98.15\% (1.2\%) & 135.16 (248.93)\\
{\small\bf Par+T\_parser}   & 93.23\% (1.1\%) & 99.11\% (0.5\%) & 94.06\% (1.5\%) & 31.17 (81.45) \\
\hline
\end{tabular}
\caption{{Means and STDs of ten experiments (OVIS): Par denotes Partial-Parser}}
\label{Tab1}
\end{table*}

Table~\ref{Tab1} shows the statistical means and  (in brackets)
the standard deviations of the ten experiments (always for sentences longer
than 1~word). 
{\bf Right parse} (also structural consistency) denotes the percentage of 
test sentences for which the parser's chart 
contains the right parse (i.e. test-set parse). {\bf Any parse} (also coverage)
denotes the percentage of test sentences for which the parser's chart contained a
parse.  {\bf Precision} denotes the ratio (Right parse/Any parse), which expresses
the precision of the parser as a parse-space generator.
And {\bf active nodes} denotes the mean number of active items in a CYK parser
implementation; active items are those items that participate in a full
parse of the sentence. 

On average the partial-parser reduces the space by 4.33 times on all sentence lengths. 
The reduction of space reaches a mean of 7 times on sentences longer than 6.
The degradation in precision (4\%) is due to several reasons.
Firstly, the fact that the partial-parser is currently
implemented as a context-free recognizer clearly contributes to this degradation.
Secondly, after analyzing the test-results of one experiment, we found out that about 
half of the errors are due to deeper structures assigned by the Partial-Parser rather
than really wrong structures; typically those were compound NPs which received shallow 
annotations in the tree-bank. 
Thirdly, part of the errors is due to tree-bank annotation mistakes.
And finally, there is a remaining part of errors which is due to the assumptions of the 
EBL method; these are harder to solve than the previous three. 

In figure~\ref{FigPlot1} and~\ref{FigPlot2} we show the learning curves of the present 
method for six sizes of training-sets; five of the six training-sets were obtained 
randomly from a set of 4500 trees, and the sixth consisted of the whole set. For these
experiments we employed the same set of 500 test-trees randomly chosen (all length sentences). 
The experiments were repeated twice: once allowing ``retreating" on local-context (as explained
earlier), and once not allowing that, during the learning phase
(the two versions are denoted ``Retreating" and ``No-Retreating" respectively).
The learning curves of the Retreating partial-parser, show that from a certain point 
on there is some deterioration of precision but further gain of space-reduction. The situation is 
different with the No-Retreating version. The explanation for the loss of precision is
that when the training-set is smaller, less PA-SSFs are learned, which implies a larger 
role for the T-Parser. This situation is magnified by the fact that the coverage of 
the T-Parser is lower on smaller training-sets. The deterioration of precision of the
Retreating version compared to the No-Retreating version is due to the fact that the 
number of learned local-context PA-SSFs becomes much larger; this implies reduction of parse-space but
also some loss of precision (since the partial-parser does not employ the local-context). 
\begin{table*}[htb]
\begin{tabular}{|c||c|c|c|c|c|c|}
\hline
                     &                  &                   & \multicolumn{3}{c|}{{\bf CPU-secs. for sentence length}}\\
{\bf System}          & {\bf Coverage}  & {\bf Accuracy}  & $\geq$ 2            & $\geq$ 7      & $\geq$ 10       \\
\hline
{\small\bf DOP}       & 95.00\% (1.4\%) & 93.50\% (0.1\%) & 3.98 (11.29)  & 13.55 (22.84) & 37.46 (41.35)\\
{\small\bf EBL+DOP}   & 94.61\% (1.4\%) & 91.72\% (0.1\%) & 1.28 (2.31)   & 2.98  (4.46)  & 6.21 (8.67)\\
{\small\bf EBL0.75+DOP} & 94.96\% (1.3\%) & 91.90\% (1.4\%) &  1.33 (2.43)   &  3.18 (4.69)  &  6.85 (8.97)\\
\hline
\hline
{\bf System}          & \multicolumn{2}{|c|}{\bf number of trees} &\multicolumn{3}{|c|}{\bf number of nodes in trees}\\
\hline
{\small\bf DOP}       & \multicolumn{2}{|c|}{27907 (1634)}  &\multicolumn{3}{|c|}{141960 (~938)}\\
{\small\bf EBL+DOP}   & \multicolumn{2}{|c|}{23660 (302~) }  &\multicolumn{3}{|c|}{117134 (1627)}\\
{\small\bf ParPar }   & \multicolumn{2}{|c|}{84.4 (4.8)} &\multicolumn{3}{|c|}{818.5 (10.26) }\\
{\small\bf EBL0.75+DOP}   & \multicolumn{2}{|c|}{23728 (138~) }  &\multicolumn{3}{|c|}{117750 (1551)}\\
{\small\bf ParPar0.75 }   & \multicolumn{2}{|c|}{80.6 (3.0)} &\multicolumn{3}{|c|}{812 (10.40) }\\
\hline
\end{tabular}
\caption{{Means and STDs of ten experiments (OVIS), ParPar denotes Partial-Parser}}
\label{Tab2}
\end{table*}
\begin{figure}
\setlength{\unitlength}{0.240900pt}
\ifx\plotpoint\undefined\newsavebox{\plotpoint}\fi
\sbox{\plotpoint}{\rule[-0.200pt]{0.400pt}{0.400pt}}%
\begin{picture}(1049,629)(80,0)
\font\gnuplot=cmr10 at 10pt
\gnuplot
\sbox{\plotpoint}{\rule[-0.200pt]{0.400pt}{0.400pt}}%
\put(220.0,113.0){\rule[-0.200pt]{184.288pt}{0.400pt}}
\put(220.0,113.0){\rule[-0.200pt]{0.400pt}{118.764pt}}
\put(220.0,187.0){\rule[-0.200pt]{4.818pt}{0.400pt}}
\put(198,187){\makebox(0,0)[r]{$75.00$}}
\put(965.0,187.0){\rule[-0.200pt]{4.818pt}{0.400pt}}
\put(220.0,261.0){\rule[-0.200pt]{4.818pt}{0.400pt}}
\put(198,261){\makebox(0,0)[r]{$150.00$}}
\put(965.0,261.0){\rule[-0.200pt]{4.818pt}{0.400pt}}
\put(220.0,335.0){\rule[-0.200pt]{4.818pt}{0.400pt}}
\put(198,335){\makebox(0,0)[r]{$225.00$}}
\put(965.0,335.0){\rule[-0.200pt]{4.818pt}{0.400pt}}
\put(220.0,409.0){\rule[-0.200pt]{4.818pt}{0.400pt}}
\put(198,409){\makebox(0,0)[r]{$300.00$}}
\put(965.0,409.0){\rule[-0.200pt]{4.818pt}{0.400pt}}
\put(220.0,483.0){\rule[-0.200pt]{4.818pt}{0.400pt}}
\put(198,483){\makebox(0,0)[r]{$375.00$}}
\put(965.0,483.0){\rule[-0.200pt]{4.818pt}{0.400pt}}
\put(220.0,557.0){\rule[-0.200pt]{4.818pt}{0.400pt}}
\put(198,557){\makebox(0,0)[r]{$450.00$}}
\put(965.0,557.0){\rule[-0.200pt]{4.818pt}{0.400pt}}
\put(373.0,113.0){\rule[-0.200pt]{0.400pt}{4.818pt}}
\put(373,68){\makebox(0,0){$10$}}
\put(373.0,586.0){\rule[-0.200pt]{0.400pt}{4.818pt}}
\put(526.0,113.0){\rule[-0.200pt]{0.400pt}{4.818pt}}
\put(526,68){\makebox(0,0){$20$}}
\put(526.0,586.0){\rule[-0.200pt]{0.400pt}{4.818pt}}
\put(679.0,113.0){\rule[-0.200pt]{0.400pt}{4.818pt}}
\put(679,68){\makebox(0,0){$30$}}
\put(679.0,586.0){\rule[-0.200pt]{0.400pt}{4.818pt}}
\put(832.0,113.0){\rule[-0.200pt]{0.400pt}{4.818pt}}
\put(832,68){\makebox(0,0){$40$}}
\put(832.0,586.0){\rule[-0.200pt]{0.400pt}{4.818pt}}
\put(985.0,113.0){\rule[-0.200pt]{0.400pt}{4.818pt}}
\put(985,68){\makebox(0,0){$50$}}
\put(985.0,586.0){\rule[-0.200pt]{0.400pt}{4.818pt}}
\put(220.0,113.0){\rule[-0.200pt]{184.288pt}{0.400pt}}
\put(985.0,113.0){\rule[-0.200pt]{0.400pt}{118.764pt}}
\put(220.0,606.0){\rule[-0.200pt]{184.288pt}{0.400pt}}
\put(100,620){\makebox(0,0){\shortstack{{\tiny\bf Number}\\{\tiny\bf of events}}}}
\put(602,23){\makebox(0,0){CPU-secs.}}
\put(220.0,113.0){\rule[-0.200pt]{0.400pt}{118.764pt}}
\put(855,541){\makebox(0,0)[r]{{\bf DOP}}}
\put(899,541){\raisebox{-.8pt}{\makebox(0,0){$\bullet$}}}
\put(373,372){\raisebox{-.8pt}{\makebox(0,0){$\bullet$}}}
\put(450,264){\raisebox{-.8pt}{\makebox(0,0){$\bullet$}}}
\put(526,227){\raisebox{-.8pt}{\makebox(0,0){$\bullet$}}}
\put(679,181){\raisebox{-.8pt}{\makebox(0,0){$\bullet$}}}
\put(985,142){\raisebox{-.8pt}{\makebox(0,0){$\bullet$}}}
\multiput(318,606)(4.749,-20.205){12}{\usebox{\plotpoint}}
\multiput(373,372)(12.049,-16.900){6}{\usebox{\plotpoint}}
\multiput(450,264)(18.661,-9.085){5}{\usebox{\plotpoint}}
\multiput(526,227)(19.877,-5.976){7}{\usebox{\plotpoint}}
\multiput(679,181)(20.589,-2.624){15}{\usebox{\plotpoint}}
\put(985,142){\usebox{\plotpoint}}
\sbox{\plotpoint}{\rule[-0.400pt]{0.800pt}{0.800pt}}%
\put(855,496){\makebox(0,0)[r]{{\bf EBL+DOP}}}
\put(899,496){\makebox(0,0){$+$}}
\put(297,209){\makebox(0,0){$+$}}
\put(373,136){\makebox(0,0){$+$}}
\put(450,123){\makebox(0,0){$+$}}
\put(526,119){\makebox(0,0){$+$}}
\put(679,117){\makebox(0,0){$+$}}
\put(985,115){\makebox(0,0){$+$}}
\sbox{\plotpoint}{\rule[-0.500pt]{1.000pt}{1.000pt}}%
\multiput(275,606)(1.148,-20.724){20}{\usebox{\plotpoint}}
\multiput(297,209)(14.969,-14.378){5}{\usebox{\plotpoint}}
\multiput(373,136)(20.466,-3.455){3}{\usebox{\plotpoint}}
\multiput(450,123)(20.727,-1.091){4}{\usebox{\plotpoint}}
\multiput(526,119)(20.754,-0.271){8}{\usebox{\plotpoint}}
\multiput(679,117)(20.755,-0.136){14}{\usebox{\plotpoint}}
\put(985,115){\usebox{\plotpoint}}
\end{picture}
\caption{Number of sentences to CPU-time} 
\label{FigPlot3}
\end{figure}
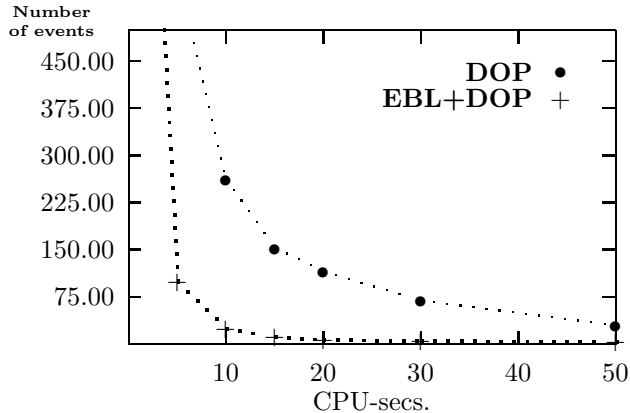
\subsection{OVIS experiments with DOP}
To test the present method together with DOP we employed the same 10 random splits
which we employed in the previous experiments. This time we did {\em not}\/ include anything 
about unknown words in the test-sets (i.e. a sentence that includes an unknown word 
is not parsable). DOP and EBL+DOP were trained employing the following parameter setting 
for partial-trees (cf.~\cite{Simaan95}): for each projected partial-tree, 
a maximum was set on its depth~(D), number of substitution-sites~(N) on its frontier, 
number of words (W) and number of consecutive words (C) on its frontier.
The setting was D=4, N=2, W=7 and C=2. This reduces the number of elementary-trees 
which DOP projects drastically without loss of accuracy.
Furthermore, the EBL algorithm was trained with a threshold on the frequency 
of SSFs equal to~14. The EBL method is used for both specializing the T-parser, which 
DOP employs prior to disambiguation~\cite{Simaan95}, and for specifying the 
cut-nodes for DOP.

Another set of experiments on the same 10 random splits (denoted EBL0.75 in table~\ref{Tab2}) 
was conducted where the threshold on $\theta$ was set at~0.75, i.e.  a sequence of grammar symbols
was allowed to be learned if it was for at least 75\% of the time an SSF. 
This was achieved by allowing the learning algorithm to change the threshold ($\theta$) on the 
definition of PA-SSF; each time there are no more PA-SSFs to learn, $\theta$ was reduced 
by 0.03 and learning went on. 

Table~\ref{Tab2} lists the means and standard deviation for the 10 experiments for
all sentences of length larger or equal to 2 words. The average (std of) percentage of the
sentences that included an unknown word is~2.56\% (0.93\%). The measures which the table lists 
are coverage and accuracy, where {\em {\bf coverage} is the percentage of sentences
that received a parse}, and {\em {\bf accuracy} is the percentage of {\it parsable}\/ 
sentences that received exactly the same parse as the test-set counterpart. 
The {\bf precision} of a method is equal to the multiplication of the two previous measures}.

On average, DOP ``guesses" in 88.82\% (i.e. precision) of 
the cases exactly the same test-set parse; with EBL this becomes 86.77\%, i.e. a loss of 2.05\%.
The speedup is on average 3.1 times but, more importantly, the standard-deviation in processing time
is less than a fifth. On longer sentences, the speedup exceeds 6 times. Figure~\ref{FigPlot3}
shows the accumulative frequency of sentences to CPU-time: for $x$~secs., the figure shows the number
of sentences that take at least $x$~secs. in parsing. If a deadline of 5~secs. is set beforehand, 
DOP misses around the 600 cases (of 3372) while the EBL misses less than 100~cases. 
At~10~secs. the figures are 263 to 23, and at~20~secs. it's 116 to 6 cases respectively.

The version EBL0.75 shows similar learning capabilities to the EBL (i.e. EBL1.0) version.
Its precision is slightly better with 87.26\% and its coverage is virtually the same as DOP's.
The EBL0.75 does not improve speedup though (actually it's slightly slower). The explanation to
this behavior is simple: EBL0.75 does not seem to learn significantly many more rules than 
EBL1.0 and, during parsing, it gives up the assumption that PA-SSF borders are trustworthy.
This way it takes less risk but then it slightly loses speed. Again we conjecture that EBL0.75 
would provide more speedup if local-context would be used during partial-parsing.
Table~\ref{Tab2} shows also the sizes of grammars which DOP projects with and without EBL.
The number of elementary-trees in the table for the Partial-Parser does not include the lexicon.
The sizes of the statistical grammars of DOP with EBL is about 1.2 times smaller than DOP's.
This is not the reduction which we hoped for, but it is quite evident that this is due to 
constraining the EBL mechanism; currently learning takes place only where local-context can be assumed 
of minor importance.

\section{Conclusions and future work}
\label{Conc}
We described a new view of EBL methods for parsing aiming directly at partial-disambiguation.
Speedup is due to fast parsing that minimizes the parse-space prior to the, often, 
expensive probabilistic disambiguation. This view is exemplified by an EBL method, 
which 1)~specializes parsers by inferring partial-parsers, and 2)~refines existing stochastic 
models of disambiguation.
%
%
From preliminary experiments with a pilot implementation we observe that the method has
the potential of speeding-up parsing, especially for Speech Understanding
where the input is a word-graph. Also we see that it is possible to minimize coverage loss
when using EBL and still gain space-reduction and speed. However, these experiments have 
shown that it is hard to gain speed and space-reduction without employing local-context 
and without extensive training-sets. 

Work on extending the parsing algorithm to accommodate local-context is being carried
out and shall be ready very soon. Further exploration will proceed on several fronts. 
We intend to test this method on larger and harder tree-banks. An implementation as
CSFTs will be studied and implemented.  We shall also study other measures of utility, 
as mentioned earlier in this paper. And finally, we might extend this method as 
in~\cite{SrinivasJoshi} or employ existing similarity-based measures for matching PA-SSFs,
instead.

\paragraph{Acknowledgements:} This work was supported by
        project~305~00~903, Priority Programme for Language and Speech Technology, Dutch 
        Organization for Scientific Research (NWO). I thank Christer Samuelsson, Remko Scha
        and Remko Bonnema for comments on earlier versions. 

\end{document}